\documentclass{llncs}
\usepackage{booktabs} 
\usepackage{comment}
\usepackage{graphicx}
\usepackage{cellspace}
\usepackage{wrapfig,picins}
\usepackage[sorting=none]{biblatex}
\usepackage{amsmath}
\usepackage{graphicx}
\usepackage{color}
\usepackage{url}
\usepackage{booktabs} 
\usepackage{subfigure}
\usepackage{amsmath}
\DeclareGraphicsExtensions{.pdf,.png,.jpg}
\usepackage[english]{babel}

\usepackage[section]{placeins}

\usepackage{listings}
\usepackage{xcolor}
\definecolor{codegreen}{rgb}{0,0.6,0}
\definecolor{codegray}{rgb}{0.5,0.5,0.5}
\definecolor{codepurple}{rgb}{0.58,0,0.82}
\definecolor{backcolour}{rgb}{0.95,0.95,0.92}
\lstdefinestyle{mystyle}{
    backgroundcolor=\color{backcolour},   
    commentstyle=\color{codegreen},
    keywordstyle=\color{magenta},
    numberstyle=\tiny\color{codegray},
    stringstyle=\color{codepurple},
    basicstyle=\ttfamily\footnotesize,
    breakatwhitespace=false,         
    breaklines=true,                 
    captionpos=b,                    
    keepspaces=true,                 
    numbers=left,                    
    numbersep=5pt,                  
    showspaces=false,                
    showstringspaces=false,
    showtabs=false,                  
    tabsize=2
}
\begin{document}

\frontmatter          
\pagestyle{empty}  

\title{Binary classification of proteins
by a Machine Learning approach}
%
%
 \author{Damiano Perri\inst{1} $^{ORCID: 0000-0001-6815-6659}$\newline Marco Simonetti\inst{1} $^{ORCID: 0000-0003-2923-5519}$\newline Andrea Lombardi\inst{2} $^{ORCID:0000-0002-7875-2697}$\newline  Noelia Faginas-Lago\inst{2} $^{ORCID:0000-0002-4056-3364}$\newline Osvaldo Gervasi\inst{3} $^{ORCID: 0000-0003-4327-520X}$ 
 }
\institute{
University of Florence, Dept. of Mathematics and Computer Science, Florence, Italy \and University of Perugia, Dept. of Chemistry, Biology and Biotechnology, Perugia, Italy\and University of Perugia, Dept. of Mathematics and Computer Science, Perugia, Italy
}
\titlerunning{Binary classification of proteins by a Machine Learning approach} 
\authorrunning{D. Perri, M.  Simonetti, A. Lombardi, M. N. Faginas Lago and O. Gervasi} 

\maketitle

\begin{abstract}
In this work we present a system based on a Deep Learning approach, by using a Convolutional Neural Network, capable of classifying protein chains of amino acids based on the protein description contained in the Protein Data Bank. Each protein is fully described in its chemical-physical-geometric properties in a file in XML format.
The aim of the work is to design a prototypical Deep Learning machinery for the collection and management of a huge amount of data and to validate it through its application to the classification of a sequences of amino acids. We envisage applying the described approach to more general classification problems in biomolecules, related to structural properties and similarities.
\end{abstract}


\section{Introduction}
Proteins in Nature exhibit a very complicate relationship between their complex structures, with considerable differences from a chemical, physical and geometric point of view and their biological functions. 
The theoretical and computational study of proteins structure and function, as well as of nucleic acids, lipid membranes and other biosystems, is mainly carried on by Molecular Dynamics (MD) simulations, grounded upon classical mechanics and Force Fields. MD is used to sample the system phase space and to capture the relevant dynamical processes of proteins across different timescales. Simulations can be carried out at different levels of details. These can be atomistic simulations, where each atom is followed in detail, or can be based on coarse grained models, where group of atoms are replaced by pseudo-atoms and a reduced number of degrees of freedom allows to model biological phenomena accessing much longer time scales.

In spite of continuous progress and increasing availability of High Performance Computing resources, many aspects in the dynamics and structure modelling still remain problematic, due to the inherent computational complexity of proteins and other biomolecules. These are related to 
(i) very high computational demand and sampling limitations and ii) limited force field accuracy, (iii) search for main stable structure and reactive or isomerization pathways, (iv) multiscale nature of dynamics.

In the last few years, the recourse to Machine Learning (ML) applications has become widespread in molecular dynamics. Particularly, approaches based on a variant of ML called deep neural networks \cite{dl1,dl2} are becoming broadly popular. These can be applied to the many classification problems that, generally speaking, occur in the theoretical and computational modelling of proteins and other biomolecules. The growing amount of both experimental and theoretical data, available in databases and repositories, makes it increasingly feasible an efficient training of neural networks.   

In this work, we have posed the basic problem of classifying as ``real'' a protein given its amino acid sequence, using a Deep Learning approach.\newline 
The choice of the network to process the sequences fell on Convolutional Neural Network (CNN). 
A choice in some ways less obvious than a Recurrent Neural Network (RNN)\cite{liu2017deep}, given the nature and form of the data to be processed, but one that has revealed to be able to return results that are quite satisfactory.
\newline 
In addition, CNNs can be easily implemented even on hardware that is not particularly performing, with the advantage of enabling to use the model developed on a plurality of platforms\cite{sze2017hardware} and even on dedicated boards\cite{zhang2015optimizing}.
The cnn are mainly used for image analysis, they allow to extract features that are then used to correctly classify objects, people or things.\cite{bocca2019}\cite{cnnDamianoDividiti2019}
In this article we explain our approach to the problem using CNN 1D, i.e. one-dimensional.
\newline The data set to train our network was taken from The Protein Data Bank (PDB) \cite{PDB}, that is a free access archive containing 3D structure data of proteins and nucleic acids.


\section{The Architecture of the system}
The system must be able to correctly classify a sequence of amino acids, telling whether it represents a ``possible'' real protein.
To get to this, we decided to use a less conventional approach, but which seemed to us very promising, based on a CNN, akind of networks particularly appreciated in the recognition of images and characteristics from two-dimensional and three-dimensional objects.

\subsection{Data extraction and processing}
Our work was developed entirely in Python working in the development environment provided by Google, called Colab.
We divided the problem into a set of sub-problems.
First of all we downloaded all the protein database provided by Protein Data Bank.
The dataset is downloadable in two different formats: PDB and XML.
Between the two formats the information content is the same, the only difference being only the data format to store with the information.
The PDB database is composed of hundreds of thousands of proteins all entirely described by single files in different formats: our choice has been placed on an open format like XML because it is much easier to be read and analyzed through the libraries made available by Python.
The first problem we had to solve was related to the size of the dataset.
In total there are 160,797 files in "xml.gz" compressed format for a total size of 57.6GB.
Once extracted the archive takes up more than 2TB of disk space due to the huge amount of data it contains.
First we created a script that solves the problem of rearranging the dataset in order to be conveniently handled by our tools. In practice, it outputs a single compressed file in which the proteins are arranged as a list. Each protein in the list is represented by a sequence of amino acids.
In this way we were able to obtain a single file named "dataset.csv.gz" with a size of 7GB, where just the relevant information was extracted, ready to be used for the next steps.
Each single protein was entirely mapped and unequivocally encoded as a very precise sequence of numbers, consisting of a dictionary of type "amino acid's name: positive integer number".\newline
A first analysis of the data showed that the amino acids present in the datased were 52, but only a part of them (23), was really present in the whole set of proteins examined, containing 105,123 instances.\newline
 In addition, the sequences had variable lengths (defined as the number of amino acids in the chain) for the various molecules, with a distribution which is shown in the table below:
\begin{table}
\centering
\begin{tabular}{@{}|c|r|r| @{}}
\toprule
\multicolumn{3}{|c|}{Length of molecules' sequences} \\ \midrule
Chain length (interval)   & Number of molecules & Number of molecules \%\\ \midrule
1 .. 9                      & 96                 & 0.09 \\ \midrule
10 .. 99                    & 3,149              & 3.00 \\ \midrule
100 .. 999                  & 83,526             & 79.46\\ \midrule
1,000 .. 1,500              & 9,107              & 8.66\\ \midrule
1,501 .. 9,999              & 9,117              & 8.67\\ \midrule
10,000 .. 99,999            & 127                & 0.12\\ \midrule
100,000 .. 1,000,000        & 1                  & 0.00\\ \bottomrule
\end{tabular}
\caption{Preliminary analysis of the dataset}
\end{table}
\newline 
Since the sequence to be fed to the network must have a predetermined fixed length, we have chosen to consider only those proteins with a length of less than 1,500, which make up about 91.21\% of the entire set of sequences.\newline
Since a protein in its constitutive unfolded sequence can be passed from left to right or vice versa, the dataset has been "augmented", by inserting the same sequences of proteins already present, but reversed. This allowed to obtain a dataset of real proteins with double size respect of the original one.\newline

\begin{figure}
    \centering
    \includegraphics[width=\linewidth]{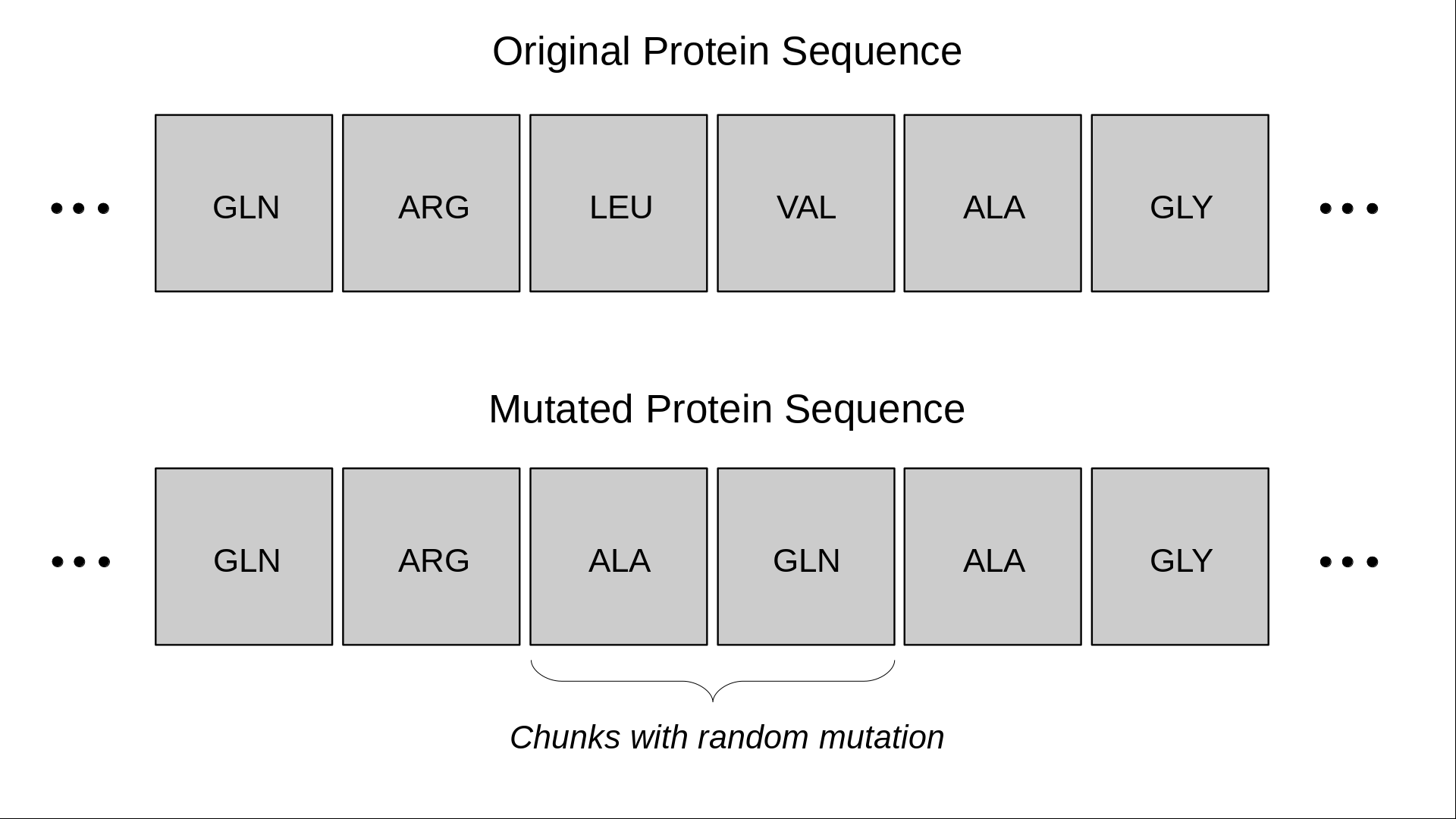}
    \caption{False chunks in sequence}
    \label{fig01}
\end{figure}

The second step for the preparation of the dataset was the generation of the fake samples (FALSE samples), in equal number to the real samples (TRUE samples).\newline
Initially, trivial cases were inserted, as whole sequences of length 1,500 of the same amino acid repeated more times, obtaining 23 new false proteins. Subsequently, the generation level became more sophisticated: in each true protein, a fragment at a random position of the protein sequence was replaced by another one, giving place to a sequence mutation; each new fragment had a random length between 5 and 7\% of the entire protein sequence length and contained amino acids randomly taken from the dictionary of all possible amino acids in our real cases (Fig.\ref{fig01}).
\subsection{Model construction and validation}
\begin{figure}
    \centering
    \includegraphics[width=\linewidth]{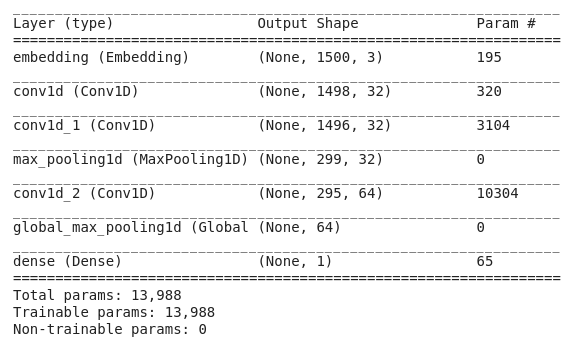}
    \caption{Structure of the network model}
    \label{fig02}
\end{figure}

\begin{figure}
    \centering
    \includegraphics[width=\linewidth]{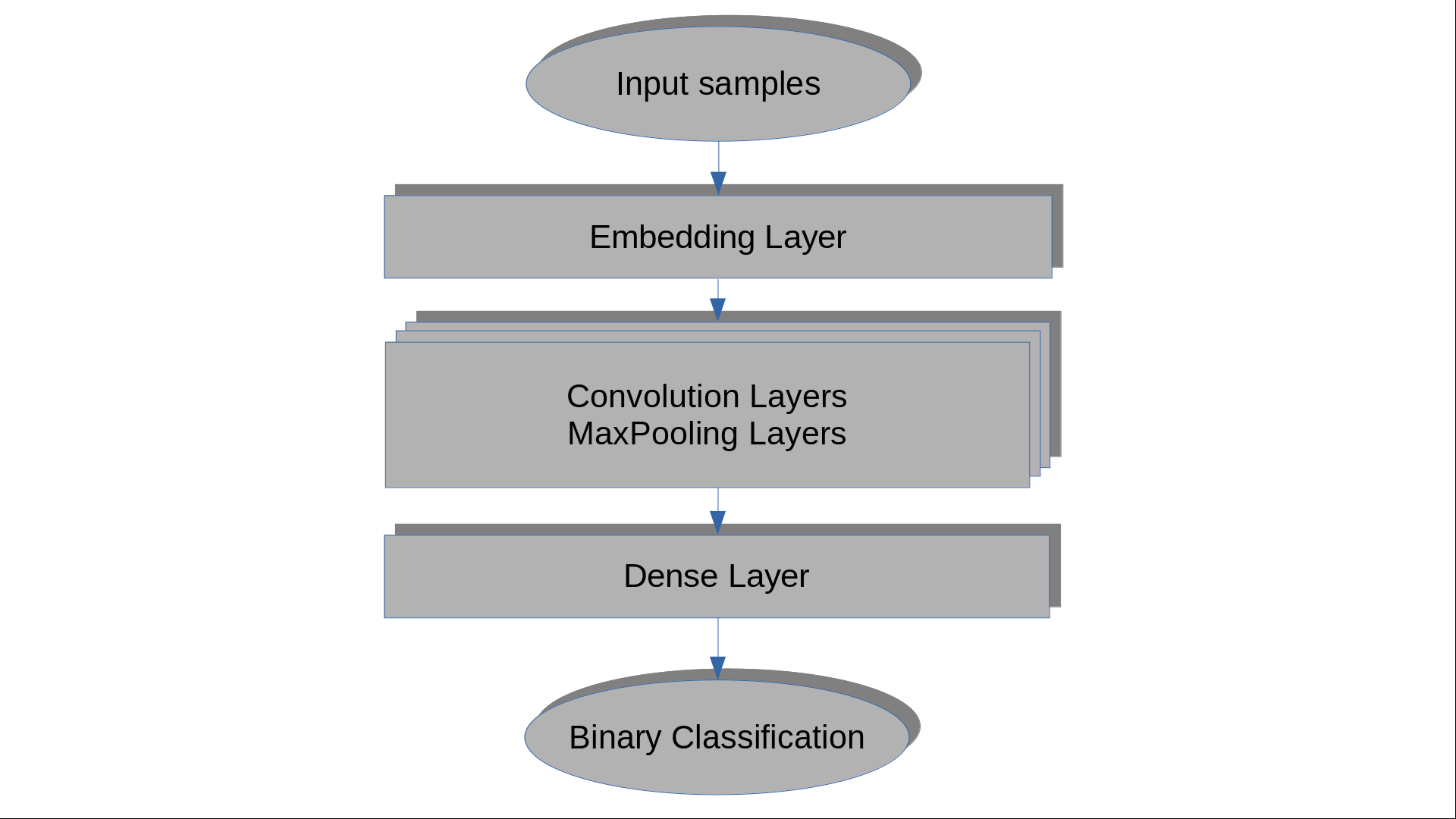}
    \caption{Flow Diagram of the network model}
    \label{fig03}
\end{figure}

During the working phases we tested several networks built with different parameters and layers. A simple model and at the same time satisfactory for the quality of the results, was the one represented in Fig.\ref{fig02} and Fig.\ref{fig03}, characterized as follows:
\begin{itemize}
    \item Embedding Layer: to create the weight matrix and to index it, according to the inputs which should be lists of positive integers (encoded amino acids)
    \item Convolution Layer: these are our first two layers that define a feature detector with a kernel equal to 3.
    \item Pooling Layer: with a size of the max pooling windows of 5.
    \item Convolution Layer: another layer that define a feature detector with a kernel equal to 5 to refine outcomes.
    \item MaxPooling Layer: to reduce the complexity of the output and prevent overfitting of the data.
    \item The final dense layer: this is a fully connected layer with SIGMOID activation; this layer will reduce the vector dimensions to a binary vector since we have 2 classes that we want to predict.
\end{itemize}
The model is extremely light, using a total of only 13,988 parameters.\newline
\newline
\newline
In the section where the model is compiled we have chosen the "binary crossentropy"\cite{rubinstein2013cross}\cite{de2005tutorial} as a feedback signal for learning the weight tensors to be minimized, as an optimizer to rule the gradient descent the "Adadelta optimizer"\cite{zeiler2012adadelta}, with a fixed learning-rate equal to 1.0 and rho equal to 0.95, and as our metric the accuracy of the model.\newline
The network was let to iterate on the training data (306,812 samples) in mini-batches of 50 samples for 50 epochs.\newline
In addition, we have inserted checkpoints to the simulation, so that we can save the best weight configuration of the model.
\subsection{Analysis of results}
\begin{figure}
    \centering
    \includegraphics[width=\linewidth]{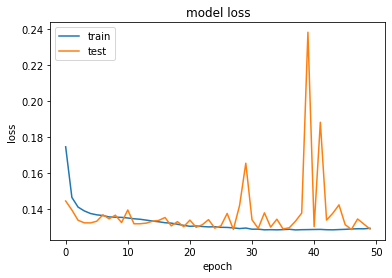}
    \caption{Loss and validation loss for the model}
    \label{fig04}
\end{figure}

\begin{figure}
    \centering
    \includegraphics[width=\linewidth]{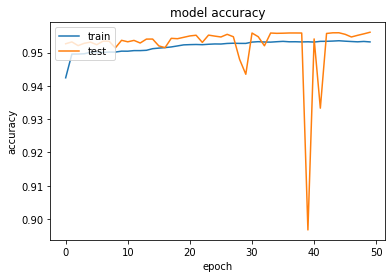}
    \caption{Accuracy and validation accuracy for the model}
    \label{fig05}
\end{figure}

\begin{figure}
    \centering
    \includegraphics[width=\linewidth]{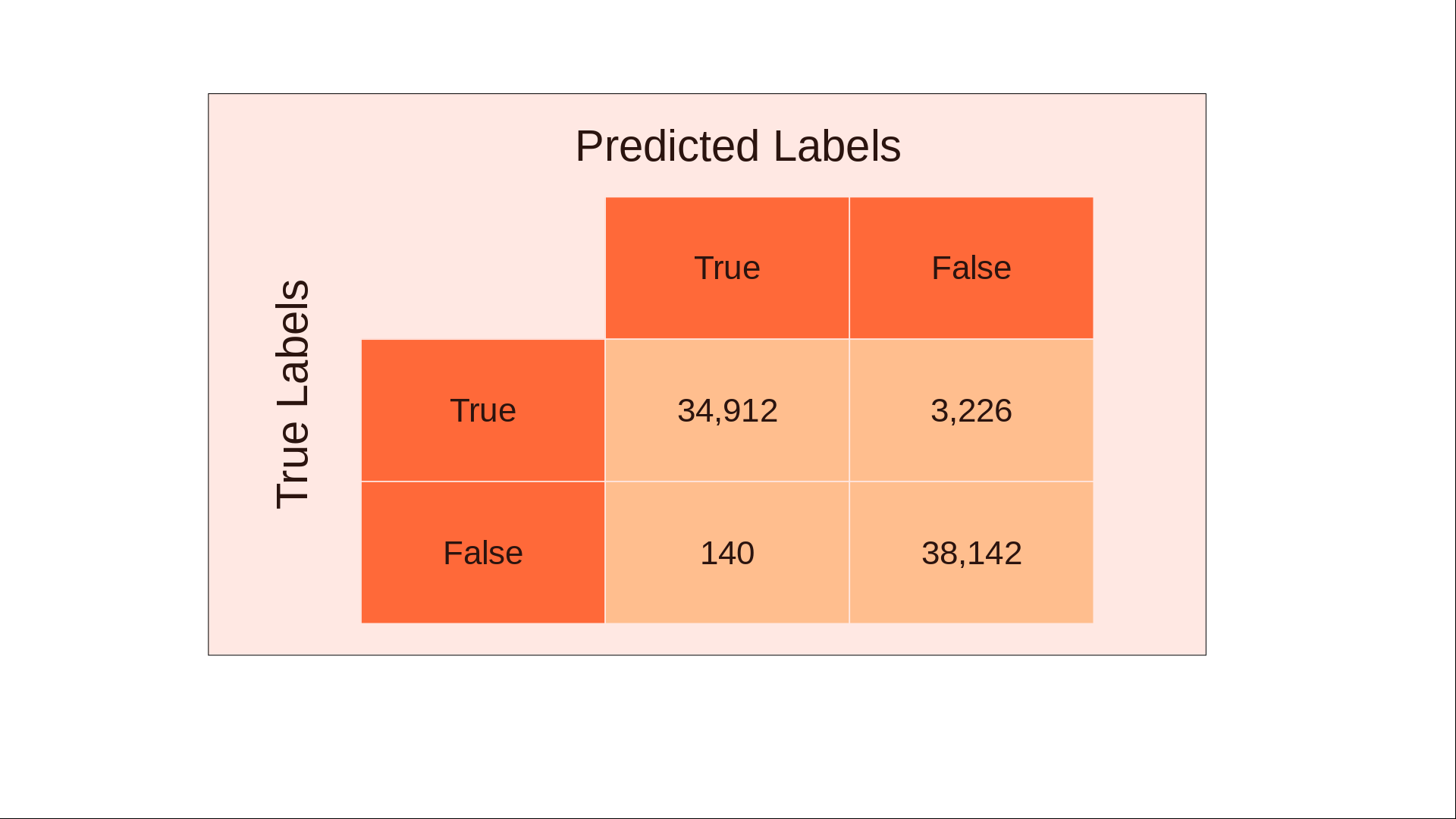}
    \caption{Confusion Matrix of results}
    \label{fig06}
\end{figure}

\begin{figure}
    \centering
    \includegraphics[width=\linewidth]{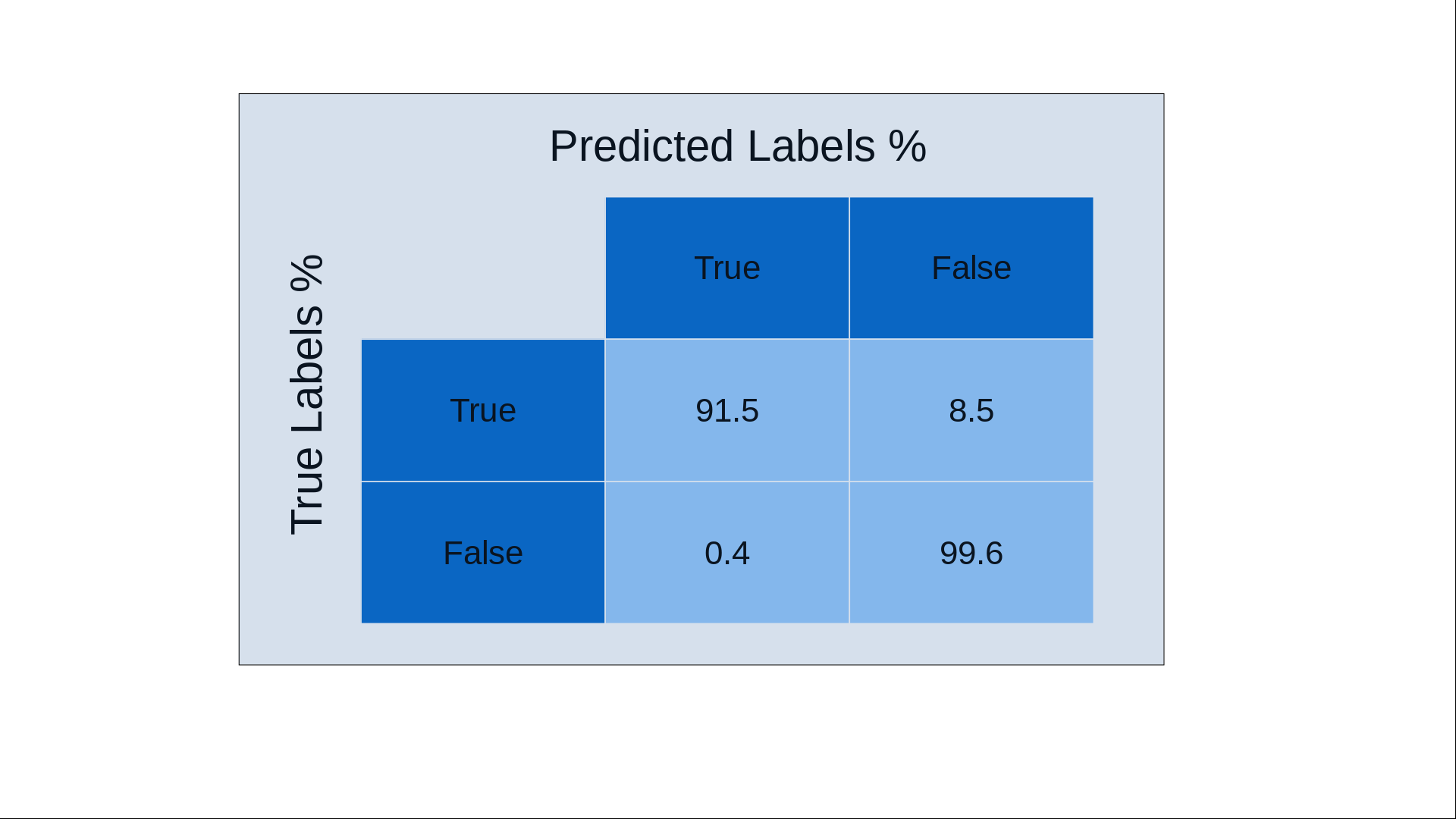}
    \caption{Confusion Matrix of results (\%)}
    \label{fig07}
\end{figure}

The results obtained are very encouraging, having achieved a score of 95.6\% on the accuracy of the tests.\newline
In Fig.\ref{fig04} and Fig.\ref{fig05} it is shown the plot the loss and accuracy of the model, as a function of the  various periods (epochs) of the simulation.
\newline Despite the presence of sudden and rare spikes in the  graphs, essentially due to a temporary displacement of the trajectory from the set of solutions by the search algorithm, it can be seen that asymptotically there is a converging attractor around which our model is stationed.\newline
The Confusion Matrix\cite{visa2011confusion} indicates that the model is well performing, while highlighting rooms for improvement, especially in the phase of choosing the optimizer and modulating the hyper-parameters, see Fig.\ref{fig06} and Fig.\ref{fig07}.\newline
The apparent high number of "true negatives" depends on the phenomenon of training data padding, necessary for neural networks of this type, which require input sequences of a predetermined length.

\section{Further development of the CNN approach to proteins as a tool for chemical modelling}

A further development of this work is to extend the Deep Learning model to the analysis of more subtle properties of the aminoacid sequences of the proteins and, above all, to classification of properties of the protein molecular structure and energy.
General possible relationships between sequence, structure and function of proteins are of particular interest.

There is no obvious way how to proceed towards the objective of feeding the CNN with different kind of protein properties, but next work clearly would imply at least an adaptation or even a reshaping of the CNN input layer, while the preparatory phase of the organization of data can be retained. 

We can now anticipate, without going into details, some of the possible ways to simply make next moves, according to some  directions as follows:\newline
\begin{itemize}
\item adding more view to the statistical analysis of sequences, recasting the learning as directed also occurrence and recurrence of single aminoacids in the protein sequences and at which position

\item adding to the sequences, here represented as a series of integer numbers, some physically meaningful parameter, such as deformation indexes \cite{xi} for the aminoacids in the chain or order parameters such as the number of hydrogen bonding interactions

\item adding to the sequences parameters describing the global minimum energy structure corresponding to the protein native status, to looking for patterns and regularities
\end{itemize}

\section{Conclusions and future works}
Despite the very promising results shown by the simulations, we believe that there is considerable room for improvement both in the selection of the system's hyper-parameters and in the preparation and management of the data needed to learn the machine.\newline
Our idea is to continue to explore the possibility of using CNN rather than RNN, as we consider it advantageous to use light models that can be easily implemented on machines with more modest hardware equipment.\newline
Furthermore, it is important to limit the use of training data padding to decrease the number of "true negatives": a working hypothesis could be the segmentation of the sequences, hypotheses on which we are currently working.

\section{Acknowledgments}

A.L. and N.F.L thank are the Dipartimento di Chimica, Biologia e Biotecnologie dell'Universit\`{a} di Perugia
(FRB, Fondo per la Ricerca di Base 2017) and the
Italian MIUR and the University of Perugia for the financial support of
the AMIS project through the program ``Dipartimenti di
Eccellenza''.
A. L. acknowledges financial support from MIUR PRIN 2015 (contract 2015F59J3R$\_$002).
A.L. thanks the OU Supercomputing Center for Education \& Research
(OSCER) at the University of Oklahoma, for allocated
computing time.
\newpage

%
%
\newpage
\printbibliography

\end{document}